# Nonparametric Bayesian Logic

**Peter Carbonetto, Jacek Kisyński, Nando de Freitas and David Poole**
Dept. of Computer Science
University of British Columbia
Vancouver, BC, Canada V6T 1Z4

## Abstract

The Bayesian Logic (BLOG) language was recently developed for defining first-order probability models over worlds with unknown numbers of objects. It handles important problems in AI, including data association and population estimation. This paper extends BLOG by adopting generative processes over function spaces — known as nonparametrics in the Bayesian literature. We introduce syntax for reasoning about arbitrary collections of objects, and their properties, in an intuitive manner. By exploiting exchangeability, distributions over unknown objects and their attributes are cast as Dirichlet processes, which resolve difficulties in model selection and inference caused by varying numbers of objects. We demonstrate these concepts with application to citation matching.

## 1 Introduction

Probabilistic first-order logic has played a prominent role in recent attempts to develop more expressive models in artificial intelligence [3, 4, 6, 8, 15, 16, 17]. Among these, the Bayesian logic (BLOG) approach [11] stands out for its ability to handle unknown numbers of objects and data association in a coherent fashion, and it does not assume unique names and domain closure.

A BLOG model specifies a probability distribution over possible worlds of a typed, first-order language. That is, it defines a probabilistic model over objects and their attributes. A model structure corresponds to a possible world, which is obtained by extending each object type and interpreting each function symbol. Objects can either be "guaranteed", meaning the extension of a type is fixed, or they can be generated from a distribution. For example, in the aircraft tracking domain [11] the times and radar blips are known, and the number of unknown aircraft may vary in possible worlds. BLOG as a case study provides a strong argument for Bayesian hierarchical methodology as a basis for probabilistic first-order logic.

BLOG specifies a prior over the number of objects. In many domains, however, it is unreasonable for the user to suggest such a proper, data-independent prior. An investigation of this issue was the seed that grew into our proposal for Nonparametric Bayesian Logic, or NP-BLOG, a language which extends the original framework developed in [11]. NP-BLOG is distinguished by its ability to handle object attributes that are generated by unbounded sets of objects. It also permits arbitrary collections of attributes drawn from unbounded sets. We extend the BLOG language by adopting Bayesian nonparametrics, which are probabilistic models with infinitely many parameters [1].

The statistics community has long stressed the need for models that avoid commiting to restrictive assumptions regarding the underlying population. Nonparametric models specify distributions over function spaces — a natural fit with Bayesian methods, since they can be incorporated as prior information and then implemented at the inference level via Bayes' theorem. In this paper, we recognize that Bayesian nonparametric methods have an important role to play in first-order probabilistic inference as well. We start with a simple example that introduces some concepts necessary to understanding the main points of the paper.

Consider a variation of the problem explored in [11]. You have just gone to the candy store and have bought a box of *Smarties* (or *M&Ms*), and you would like to discover how many colours there are (while avoiding the temptation to eat them!). Even though there is an infinite number of colours to choose from, the candies are coloured from a finite set. Due to the manufacturing process, Smarties may be slightly discoloured. You would like to discover the unknown (true) set of colours by randomly picking Smarties from the box and observing their colours. After a certain number of draws, you would like to answer questions such as: How many different colours are in the box? Do two Smarties have the same colour? What is the probability that the first candy you select from a new box is a colour you have never seen before?

The graphical representation of the BLOG model is shown in Fig. 1a. The number of Smarties of different colours, $n(\mathsf{Smartie})$, is chosen from a Poisson distribution with mean $\gamma_{\text{Smartie}}$. A colour for each Smartie $s$ is drawn from the distribution $H_{\text{ColourDist}}$. Then, for every draw $d$, $z_{\text{SmartieDrawn}}[d]$ is drawn uniformly from the set of Smarties $\{1, \ldots, n(\text{Smartie})\}$. Finally, we sample the observed, noisy colour of each draw conditioned on $z_{\text{SmartieDrawn}}[d]$ and the true colours of the Smarties.

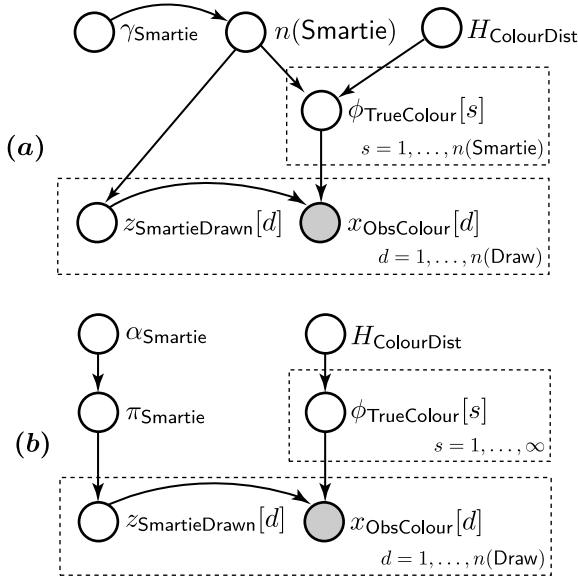

Figure 1: *(a)* The BLOG and *(b)* NP-BLOG graphical models for counting Smarties. The latter implements a Dirichlet process mixture. The shaded nodes are observations.

The NP-BLOG model for the same setting is shown in Fig. 1b. The true colours of an infinite sequence of Smarties $s$ are sampled from $H_{\text{ColourDist}}$. $\pi_{\text{Smartie}}$ is a distribution over the choice of coloured Smarties, and is sampled from a uniform Dirichlet distribution with parameter $\alpha_{\text{Smartie}}$. Once the Smarties and their colours are generated, the true Smartie for draw $d$, represented by the indicator $z_{\text{SmartieDrawn}}[d] = s$, is sampled from the distribution of Smarties $\pi_{\text{Smartie}}$. The last step is to sample the observed colour, which remains the same as in the BLOG model.

One advantage of the NP-BLOG model is that it determines a posterior over the number of Smarties colours without having to specify a prior over $n(\text{Smartie})$. This is important since this prior is difficult to specify in many domains. A more significant advantage is that NP-BLOG explicitly models a distribution over the collection of Smarties. This is not an improvement in expressiveness — one can always reverse engineer a parametric model given a target nonparametric model in a specific setting. Rather, nonparametrics facilitate the resolution of queries on unbounded sets, such as the colours of Smarties. This plays a key role in making inference tractable in sophisticated models with object properties that are themselves unbounded collections of objects. This is the case with the citation matching model in Sec. 3.1, in which publications have collections of authors.

The skeptic might still say, despite these advantages, that it is unreasonable to expect a domain expert to implement nonparametrics considering the degree of effort required to grasp these abstract notions. We show that Bayesian nonparametrics lead to sophisticated representations that can be *easier* to implement than their parametric counterparts. We formulate a language that allows one to specify nonparametric models in an intuitive manner, while hiding complicated implementation details from the user. Sec. 3 formalizes our proposed language extension as a set of rules that map code to a nonparametric generative process. We emphasize that NP-BLOG is an extension to the BLOG language, so it retains all the functionality specified in [11].

We focus on an important class of nonparametric methods, the Dirichlet process (DP), because it handles distributions over unbounded sets of objects as long as the objects themselves are infinitely exchangeable, a notion formalized in Sec. 3.4. The nonparametric nature of DPs makes them suitable for solving model selection problems that arise in the face of identity uncertainty and unknown numbers of objects. Understanding the Dirichlet process is integral to understanding NP-BLOG, so we devote a section to it. Sec. 3.5 shows how DPs can characterize collections of objects. Models based on DPs have been shown to be capable of solving a variety of difficult tasks, such as topic-document retrieval [2, 21]. Provided the necessary expert knowledge, our approach can attack these applications, and others. We conduct a citation matching experiment in Sec. 4, demonstrating accurate and efficient probabilistic inference in a real-world problem.

## 2 Dirichlet processes

A Dirichlet process $G \mid \alpha, H \sim DP(\alpha, H)$, with parameter $\alpha$ and base measure $H$, is the unique probability measure defined $G$ on the space of all probability measures $(\Phi, \mathcal{B})$, where $\Phi$ is the sample space, satisfying

$$(G(\beta_1), ..., G(\beta_K)) \sim \text{Dirichlet}(\alpha H(\beta_1), ..., \alpha H(\beta_K)) \quad (1)$$

for every measurable partition $\beta_1, \ldots, \beta_K$ of $\Phi$. The base measure $H$ defines the expectation of each partition, and $\alpha$ is a precision parameter. One can consider the DP as a generalization of the Dirichlet distribution to infinite spaces.

In Sec. 3.4, we formalize exchangeability of unknown objects. In order to explain the connection between exchangeability and the DP, it is instructive to construct DPs with the Pólya urn scheme [5]. Consider an urn with balls of $K$ possible colours, in which the probability of the first ball being colour $k$ is given by $H_k$. We draw a ball from the urn, observe its colour $\phi_1$, then return it to the urn along with another ball of the same colour. We then make another draw, observing its colour with probability $p(\phi_2 = k \mid \phi_1) = (\alpha H_k + \delta(\alpha_1 = k))/(\alpha + 1)$. After $N$ observations, the colour of the next ball is distributed as

$$P(\phi_{N+1} = k \mid \phi_{1:N}) = \frac{\alpha H_k}{\alpha + N} + \frac{\sum_{i=1}^{N} \delta(\phi_i = k)}{\alpha + N}. \quad (2)$$

The marginal $P(\phi_{1:N})$ of this process, obtained by applying the chain rule to successive predictive distributions, can be shown to satisfy the infinite mixture representation

$$P(\phi_{1:N}) = \int_{\mathcal{M}(\Phi)} \left( \prod_{k=1}^{K} \pi_k^{\sum_{i=1}^{N} \delta(\phi_i=k)} \right) DP_{\alpha,H}(d\pi), \quad (3)$$

where the $\pi_k$ are multinomial success rates of each colour $k$. This result, a manifestation of de Finetti's theorem, establishes the existence and uniqueness of the DP prior for the Pólya urn scheme [5]. In the Pólya urn setting, observations $\phi_i$ are infinitely exchangeable and independently distributed given the measure $G$. Thus, what we have established here in a somewhat cursory fashion is the appropriateness of the DP for the case when the observations $\phi_i$ are infinitely exchangeable.

Analogously, if the urn allows for infinitely many colours, then for any measurable interval $\beta$ of $\Phi$ we have

$$p(\phi_{N+1} \in \beta | \phi_{1:N}) = \frac{\alpha H(\beta)}{\alpha + N} + \frac{1}{\alpha + N} \sum_{i=1}^{N} \delta(\phi_i \in \beta).$$

The first term in this expansion corresponds to prior knowledge and the second term corresponds to the empirical distribution. Larger values of $\alpha$ indicate more confidence in the prior $H$. Note that, as $N$ increases, most of the colours will be repeated. Asymptotically, one ends up sampling colours from a possibly large but finite set of colours, achieving a clustering effect. Nonetheless, there is always some probability of generating a new cluster.

DPs are essential building blocks in our formulation of nonparametric first-order logic. In the literature, these blocks are used to construct more flexible models, such as DP mixtures and hierarchical or nested DPs [2, 21]. Since observations are provably discrete, DP mixtures add an additional layer $x_i \sim P(x_i|\phi_i)$ in order to model continuous draws $x_i$ from discrete mixture components $\phi_i$.

In the Pólya urn scheme, $G$ is integrated out and the $\phi_i$'s are sampled directly from $H$. Most algorithms for sampling DPs are based on this scheme [2, 13, 21]. In the hierarchies constructed by our language, however, we rely on an explicit representation of the measure $G$ since it is not clear we can always integrate it out, even when the measures are conjugate. This compels us to use the stick-breaking construction [19], which establishes that *i.i.d.* sequences $w_k \sim \text{Beta}(1, \alpha)$ and $\phi_k \sim H$ can be used to construct the equivalent empirical distribution $G = \sum_{k=1}^{\infty} \pi_k \delta(\phi_k)$, where the stick-breaking weights $\pi_k = w_k \prod_{j=1}^{k-1}(1 - w_j)$ and can be shown to sum to unity. We abbreviate the sampling of the weights as $\pi_k \sim \text{Stick}(\alpha)$. This shows that $G$ is an infinite sum of discrete values. The DP mixture due to the stick-breaking construction is

$$\begin{aligned} \phi_i \,|\, H &\sim H & \pi \,|\, \alpha &\sim \text{Stick}(\alpha) \\ z_i \,|\, \pi &\sim \pi & x_i \,|\, \phi_i, z_i &\sim p(x_i|\phi_{z_i}), \end{aligned} \quad (4)$$

where $z_i = k$ indicates that sample $x_i$ belongs to component $k$. The Smarties model (Fig. 1b) is in fact an example of a DP mixture, where the unbounded set of colours is $\Phi$. By grounding on the support of the observations, the true number of colours $K$ is finite. At the same time, the DP mixture is open about seeing new colours as new Smarties are drawn. In NP-BLOG, the unknown objects are the mixture components.

NP-BLOG semantics (Sec. 3) define arbitrary hierarchies of Dirichlet process mixtures. By the stick-breaking construction (4), every random variable $x_i$ has a countable set of ancestors (the unknown objects), hence DP mixtures preserve the well-definedness of BLOG models.

To infer the hidden variables of our models, we employ the efficient blocked Gibbs sampling algorithm developed in [7] as one of the steps in the overall Gibbs sampler. One complication in inference stems from the fact that a product of Dirichlets is difficult to simulate. Teh *et al.* [21] provide a solution using an auxiliary variable sampling scheme.

## 3 Syntax and semantics

This section formalizes the NP-BLOG language by specifying a procedure that takes a set of statements $\mathcal{L}_\Psi$ in the language and returns a model $\Psi$. A model comprises a set of types, function symbols, and a distribution over possible worlds $\omega \in \Omega_\Psi$. We underline that our language retains all the functionality of BLOG. Unknown objects must be infinitely exchangeable, but this trivially the case in BLOG. Sec. 3.4 elaborates on this.

We illustrate the concepts introduced in this section with an application to citation matching. Even though our citation matching model doesn't touch upon all the interesting aspects of NP-BLOG, the reader will hopefully find it instrumental in understanding the semantics.

### 3.1 Citation matching

One of the main challenges in developing an automated citation matching system is the resolution of identity uncertainty: for each citation, we would like to recover its true title and authors. For instance, the following citations from the CiteSeer database probably refer to the same paper:

Kozierok, Robin, and Maes, Pattie, A Learning Interface Agent for Meeting Scheduling, Proceedings of the 1993 International Workshop on Intelligent user Interfaces, ACM Press, NY.

R. Kozierok and P. Maes. A learning interface agent for scheduling meetings. In W. D. Gray, W. E. Heey, and D. Murray, editors, Proc. of the Internation al Workshop on Intelligent User Interfaces, Orlando FL, New York, 1993. ACM Press.

Even after assuming the title and author strings have been segmented into separate fields (an open research problem itself!), citation matching still exhibits serious challenges: two different strings may refer to the same author (e.g. J.F.G. de Freitas and Nando de Freitas) and, conversely, the same string may refer to different authors (e.g. David Lowe in vision and David Lowe in quantum field theory).

```
01 type Author; type Pub; type Citation;
02 guaranteed Citation;
03 #Author ∼ NumAuthorsDist();
04 #Pub ∼ NumPubsDist();
05 Name(a) ∼ NameDist();
06 Title(p) ∼ TitleDist();
07 NumAuthors(p) ∼ NumAuthorsDist();
08 RefAuthor(p, i) if Less(i, NumAuthors(p))
     then ∼ Uniform(Author a);
09 RefPub(c) ∼ Uniform(Pub p);
10 CitedTitle(c) ∼ TitleStrDist(Title(RefPub(c)));
11 CitedName(c, i) if Less(i, NumAuthors(RefPub(c)))
     then ∼ NameStrDist(Name(RefAuthor(RefPub(c), i)));
```

Figure 2: BLOG model for citation matching [10].

```
01 type Author; type Pub;
02 type Citation; type AuthorMention;
03 guaranteed Citation; guaranteed AuthorMention;
04 Name(a) ∼ NameDist{};
05 Title(p) ∼ TitleDist{};
06 CitedTitle(c) ∼ TitleStrDist{Title(RefPub(c))};
07 RefAuthor(u) ∼ PubAuthorsDist(RefPub(CitedIn(u)));
08 CitedName(u) ∼ NameStrDist{Name(RefAuthor(u))};
```

Figure 3: NP-BLOG model for citation matching.

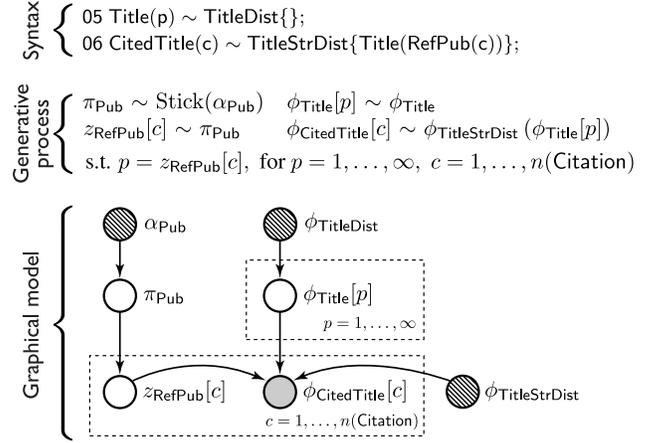

Figure 4: Three representations of lines 5-6 in Fig. 3: as an NP-BLOG program, as a generative process, and as a graphical model. Darker, hatched nodes are fixed or generated from other lines and shaded nodes are observed. Note the similarity between the graphical model and Fig. 1b. Lines 5-6 describe a DP mixture (4) over the publications $p$, where the base measure is $\phi_{\text{TitleDist}}$, $\pi_{\text{Title}}$ is the hidden distribution over publication objects, the indicators are the true publications $z_{\text{RefPub}}[c]$ corresponding to the citations $c$, and the continuous observations are the titles $\phi_{\text{CitedTitle}}[c]$.

There are a number of different approaches to this problem. Pasula et al. incorporate unknown objects and identity uncertainty into a probabilistic relational model [14]. Wellner et al. resolve identity uncertainty by computing the optimal graph partition in a conditional random field [22]. We elaborate on the BLOG model presented in [10] in order to contrast it with the one we propose. The BLOG model is shown in Fig. 2 with cosmetic modifications and the function declaration statements omitted.

The BLOG model describes a generative sampling process. Line 1 declares the object types, and line 2 declares that the citations are guaranteed (hence are not generated by a number statement). Lines 3 and 4 are number statements, and lines 5-11 are dependency statements; their combination defines a generative process. The process starts by choosing a certain number of authors and publications from their respective prior distributions. Then it samples author names, publication titles and the number of authors per publication. For each author string i in a citation, we choose the referring author from the set of authors. Finally, the properties of the citation objects are chosen. For example, generating an interpretation of CitedTitle(c) for citation c requires values for RefPub(c) and estimates of publication titles. TitleStrDist(s) can be interpreted as a measure that adds noise in the form of perturbations to input string s.

The NP-BLOG model in Fig. 3 follows a similar generative approach, the key differences being that it samples collections of unknown objects from DPs, and it allows for uncertainty in the order of authors in publications. But what do we gain by implementing nonparametrics? The advantage lies in the ability to capture sophisticated models of unbounded sets of objects in a high-level fashion, and the relative ease of conducting inference, since nonparametrics can deal gracefully with the problem of model selection.

One can view a model such as the automatic citation matcher from three perspectives: it is a set of statements in the language that comprise a program; from a statistician's point of view, the model is a process that samples the defined random variables; and from the perspective of machine learning, it is a graphical model. Fig. 3 interprets lines 5-6 of Fig. 4 in three different ways. The semantics, as we will see, formally unify all three perspectives.

Both BLOG and NP-BLOG can answer the following queries: Is the referring publication of citation $c$ the same as the referring publication of citation $d$? How many authors are there in the given citation database? What are the names of the authors of the publication referenced by citation $c$? How many publications contain the author $a$, where $a$ is one of the authors in the publication referenced by citation $c$? And what are the titles of those publications? However, only NP-BLOG can easily answer the following query: what group of researchers do we expect to be authors in a future, unseen publication?

### 3.2 Objects and function symbols

This section is largely devoted to defining notation so that we can properly elaborate on NP-BLOG semantics in Sections 3.3 to 3.5. The notation as it appears in these sections makes the connection with both first-order logic and the Dirichlet process mixture presented in Sec. 2.

The set of objects of a type $\tau$ is called the *extension* of $\tau$, and is denoted by $[\tau]$. In BLOG, extensions associated with unknown (non-guaranteed) types can vary over possible worlds $\omega$, so we sometimes write $[\tau]^\omega$. The size of $[\tau]^\omega$ is denoted by $n^\omega(\tau)$.[1] Note that objects may be unknown even if there is a fixed number of them. Guaranteed objects are present in all possible worlds. We denote $\Omega_\Psi$ to be the set of possible worlds for model $\Psi$.

A model introduces a set of function symbols indexed by the objects. For conciseness, we treat predicates as Boolean functions and constants as zero-ary functions. For example, the citation matching model (Fig. 3) has the function symbols Name and CitedTitle, among others, so there is a Name($a$) for every author $a$ and CitedTitle($c$) for every citation $c$. By assigning numbers to objects as they are generated, we can consider logical variables $a$ and $c$ to be indices on the set of natural numbers. Since BLOG is a typed language, the range of interpretations of a function symbol $f$ is specified by its type signature. For example, the interpretation of RefAuthor($u$), for each $u \in [\text{AuthorMention}] = \{1, 2, \ldots, n(\text{AuthorMention})\}$, takes a value on the range [Author]. Likewise, Title($p$) ranges over the set of strings [String]. Figures 2 and 3 omit function declaration statements, which specify type signatures. Nonetheless, this should not prevent the reader from deducing the type signatures of the functions via the statements that generate them.

Nonparametric priors define distributions over probability measures, so we need function symbols that uniformly refer to them. Letting $\mathcal{X}$ and $\mathcal{Y}$ be object domains (e.g. $\mathcal{X} = [\text{Author}]$), we define $\mathcal{M}_\mathcal{D}(\mathcal{X} \mid \mathcal{Y})$ to be the set of conditional probability densities $p(x \in \mathcal{X} \mid y \in \mathcal{Y})$ following the class of parameterizations $\mathcal{D}$. We can extend this logic, denoting $\mathcal{M}_{\mathcal{D}'}(\mathcal{M}_\mathcal{D}(\mathcal{X} \mid \mathcal{Y}) \mid \mathcal{Z})$ to be the set of probability measures $p(d \in \mathcal{D} \mid z \in \mathcal{Z})$ over the choice of parameterizations $d \in \mathcal{D}$, conditioned on $\mathcal{Z}$. And so on. For peace of mind, we assume each class of distributions $\mathcal{D}$ is defined on a measurable $\sigma$-field and the densities are integrable over the range of the sample space. Note that $\mathcal{Y}$ or $\mathcal{Z}$, but not $\mathcal{X}$, may be a Cartesian product over sets of objects. BLOG does not allow return types that are tuples of objects, so we restrict distributions of objects accordingly. One can extend the above reasoning to accommodate distributions over multiple unknown objects by adopting slightly more general notation involving products of sets of objects.

We assign symbols to all the functions defined in the language $\mathcal{L}_\Psi$. For instance, the range of NameDist in Fig. 3 is $\mathcal{M}([\text{String}])$ for some specified parameterization class. Since NameDist is not generated in another line, it must be fixed over all possible worlds. For each publication $p$, the interpretation of symbol PubAuthorsDist($p$) is assigned a value on the space $\mathcal{M}_{\text{Multinomial}}([\text{Author}])$. That is, the function symbol refers to a *distribution* over author objects. How the model chooses the success rate parameters for this multinomial distribution, given that it is not on the left side of any generating statement, is the subject of Sec. 3.5.

NP-BLOG integrates first-order logic with Bayesian nonparametric methods, but we have left out one piece of the puzzle: how to specify distributions such as NameDist, or classes of distributions. This is an important design decision, but an implementation level detail, so we postpone it to future work. For the time being, one can think of parameterizations as object classes in a programming language such as Java that generate samples of the appropriate type. We point out that there is already an established language for constructing hierarchical Bayesian models, BUGS [20].

The truth of any first-order sentence is determined by a possible world in the corresponding language. A possible world $\omega \in \Omega_\Psi$ consists of an extension $[\tau]^\omega$ for each type $\tau$ and an interpretation for each function symbol $f$. Sec. 3.5 details how NP-BLOG specifies a distribution over $\Omega_\Psi$.

### 3.3 Dependency statements for known objects

The dependency statement is the key ingredient in the specification of a generative process. We have already seen several examples of dependency statements, and we formalize them here. It is well explained in [11], but we need to extend the definition in the context of nonparametrics.

In BLOG, a dependency statement looks like

$$f(x_1, \ldots, x_L) \sim g(t_1, \ldots, t_N); \qquad (5)$$

where $f$ is a function symbol, $x_1, \ldots, x_L$ is a tuple of logical variables representing arguments to the function, $g$ is a probability density conditioned on the arguments $t_1, \ldots, t_N$, which are terms or formulae in the language $\mathcal{L}_\Psi$ in which the logical variables $x_1, \ldots, x_L$ may appear. The dependency statement carries out a generative process. For an example, let's look at the dependency statement on line 10 of Fig. 2. Following the rules of semantics [11], line 10 generates assignments for random variables $\phi_{\text{CitedTitle}}[c]$, for $c = 1, \ldots, n(\text{Citation})$, from probability density $g$ conditioned on values for $z_{\text{RefPub}}[c]$ and $\phi_{\text{Title}}[p]$, for all $p = 1, \ldots, n(\text{Pub})$. As in [11], we use square brackets to index random variables, instead of the statistics convention of using subscripts.

In NP-BLOG, the probability density $g$ is itself a function symbol, and the dependency statement is given by

$$f(x_1, \ldots, x_L) \sim g(t_1, \ldots, t_M)\{t_{M+1}, \ldots, t_{M+N}\}; \qquad (6)$$

where $f$ and $g$ are function symbols, and $t_1, \ldots, t_{M+N}$ are formulae of the language as in (5). For this to be a valid statement, $g(t_1, \ldots, t_M)$ must be defined on the range $\mathcal{M}(\mathcal{X} \mid \mathcal{Y})$, where $\mathcal{X}$ is the range of $f(x_1, \ldots, x_L)$ and $\mathcal{Y}$ is the domain of the input arguments within the curly braces. The first $M$ terms inside the parentheses are evaluated in possible world $\omega$, and their resulting values determine the choice of measure $g$. The terms inside the curly braces are evaluated in $\omega$ and the resulting values are passed to distribution $g(t_1, \ldots, t_M)$. When all the logical variables $x_1, \ldots, x_L$ refer to guaranteed objects, the semantics of the dependency statement are given by [11]. The curly brace notation is used to disambiguate the two roles of input argument variables. The arguments inside parentheses are indices to function symbols (e.g. the c in RefPub(c) in Fig. 3), whereas the arguments inside curly braces serve as input to probability densities (e.g. the term inside the curly braces in TitleStrDist{Title(RefPub(c))}). This new notation is necessary when a distribution takes both types of arguments. We don't have such an example in citation matching, so we borrow one from an NP-BLOG model in the aircraft tracking domain:[2]

---

[1] Even though the DP imposes a distribution over an infinite set of unknown objects, $n^\omega(\tau)$ is still finite since it refers to the estimated number of objects in $\omega$. $n(\tau)$ corresponds to the random variables of the DP mixture, as explained in Sec. 3.5.

> State(a, t) if t = 0 then ∼ InitState{}
> else ∼ StateTransDist(a){State(a, t − 1)};

The state of the aircraft $a$ at time $t$ is an R6Vector object which stores the aircraft's position and velocity in space. When $t > 0$, the state is generated from the transition distribution of aircraft $a$ given the state at the previous time step. StateTransDist($a$) corresponds to a measure on the space $\mathcal{M}([\text{R6Vector}] \mid [\text{R6Vector}])$.

For example, in line 6 of Fig. 3, the citation objects are guaranteed. Following the rules of semantics, line 6 defines a random variable $\phi_{\text{CitedTitle}}[c]$ corresponding to the interpretation of function symbol CitedTitle($c$) for every citation $c$. Given assignments to $\phi_{\text{TitleStrDist}}$, $z_{\text{RefPub}}[c]$ (we use $z$ to be consistent with the notation of the semantics used in this paper, although it makes no difference in BLOG) and $\phi_{\text{Title}}[p]$ for all $p \in [\text{Pub}]$ — assignments that are either observed or generated from other statements — the dependency statement defines the generative process

$$\phi_{\text{CitedTitle}}[c] \sim \phi_{\text{TitleStrDist}}(\phi_{\text{Title}}[p]) \text{ s.t. } p = z_{\text{RefPub}}[c].$$

BLOG allows for contingencies in dependency statements. These can be subsumed within our formal framework by defining a new measure $\phi'(\boldsymbol{c}, \boldsymbol{t}) = \sum_i \delta(c_i) \phi_i(t_{i,1}, t_{i,2}, \ldots)$, where $\delta(\cdot)$ is the indicator function, $c_i$ is the condition $i$ which must be satisfied in order to sample from the density $\phi_i$, $\boldsymbol{c}$ and $\boldsymbol{t}$ are the complete sets of terms and conditions, and the summation is over the number of clauses. Infinite contingencies and their connection to graphical models are discussed in [12].

### 3.4 Exchangeability and unknown objects

Unknown objects are precisely those which are not guaranteed. In this section, we formalize some important properties of generated objects in BLOG. We adopt the notion of exchangeability [1] to objects in probabilistic first-order logic. We start with some standard definitions.

---
[2] In which aircraft in flight appear as blips on a radar screen, and the objectives are to infer the number of aircraft and their flight paths and to resolve identity uncertainty, arising because a blip might not represent any aircraft or, conversely, an aircraft might produce multiple detections [10].

**Definition 1.** *The random variables $x_1, \ldots, x_N$ are (finitely) exchangeable under probability density function $p$ if $p$ satisfies $p(x_1, \ldots, x_N) = p(x_{\pi(1)}, \ldots, x_{\pi(N)})$ for all permutations $\pi$ on $\{1, \ldots, N\}$* [1].

When $n$ is finite, the concept of exchangeability is intuitive: the ordering is irrelevant since possible worlds are equally likely. The next definition extends exchangeability to unbounded sequences of random variables.

**Definition 2.** *The random variables $x_1, x_2, \ldots$ are infinitely exchangeable if every finite subset is finitely exchangeable* [1].

Exchangeability is useful for reasoning about distributions over properties on sets of objects in BLOG. From Definitions 1 and 2, we have the following result.

**Proposition 1.** *It is possible to define $g$ in the dependency statements (5) and (6) such that the sequence of objects $x_1, \ldots, x_L$ is finitely exchangeable if and only if the terms $t_1, \ldots, t_{M+N}$ do not contain any statements referring to a particular $x_l$.*

For example, the distribution of hair colours of two people, Eric and Mike, is not exchangeable given evidence that Eric is the father of Mike. What about sequences of objects such as time? As long as we do not set the predecessor function beforehand, any sequence is legally exchangeable.

In this paper, models are restricted to *infinitely* exchangeable unknown objects. We can interpret this presupposition this way: if we reorder a sequence of objects, then their probability remains the same. If we add another object to the sequence at some arbitrary position, both the original and new sequence with one more object are exchangeable. We can then appeal to de Finetti's theorem (3), and hence the Dirichlet process. Therefore, the order of unknown objects is not important, and we can reason about set of objects rather than sequences. While there are many domains in which one would like to infer the presence of objects that are not infinitely exchangeable, this constraint leaves us open to modeling a wide range of interesting domains.

Unknown or non-guaranteed objects are assigned *non-rigid designators*; a symbol in different possible worlds does not necessarily refer to the same object, and so it does not make sense to assign it a rigid label. This consideration imposes a constraint: we can only refer to a publication $p$ via a guaranteed object, such as a citation $c$ that refers to it. While we cannot form a query that addresses a specific unknown object, or a subset of unknown objects, we can pose questions about publications using existential and universal quantifiers (resolved using Skolemization, for instance). We could ask, for instance, how many publications have three or more authors.

### 3.5 Dependency statements for unknown objects

Sec. 3.2 formalized the notion of type extensions and function symbols in NP-BLOG programs. Sec. 3.3 served up the preliminaries of syntax and semantics in dependency statements. The remaining step to complete the full prescription of the semantics as a mapping from the language $\mathcal{L}_\Psi$ to a distribution over possible worlds. This is accomplished by constructing a Bayesian hierarchical model over random variables $\{\phi, n, \gamma\}$, such that the set of random variables $\phi$ is in one-to-one correspondence with the set of function interpretations, $n$ refers to the sizes of the type extensions, and $\gamma$ is a set of auxiliary random variables such that $\int p(\phi, n, \gamma) d\gamma = p(\phi, n)$. One might wonder why we don't dispense of function symbols entirely and instead describe everything using random variables, as in [18]. The principal reason is to establish the connection with first-order logic. Also, we want to make it clear that some random variables do not map to any individual in the domain. What follows is a *procedural* definition of the semantics. We now define distributions over the random variables, and their mapping to the symbols of the first-order logic.

In order to define the rules of semantics, we collect dependency and number statements according to their input argument types. If the collection of statements includes a number statement, then the rules of semantics are given in [11]. Otherwise, we describe how the objects and their properties are implicitly drawn from a DP. Consider a set of $K$ dependency statements such that the generated functions $f_1, \ldots, f_K$ all require a single input of type $v$, and $[v]^\omega$ can vary over possible worlds $\omega$. We denote $x$ to be the logical variable that ranges over $[v]$. (The output types of the $f_k$'s are not important.) The $K$ dependency statements look like

$$f_1(x) \sim g_1(t_{1,1}, \ldots, t_{1,M_1})\{t_{1,M_1+1}, \ldots, t_{1,M_1+N_1}\};$$
$$\vdots \qquad \vdots \qquad (7)$$
$$f_K(x) \sim g_K(t_{K,1}, \ldots, t_{K,M_K})\{t_{K,M_K+1}, \ldots, t_{K,M_K+N_K}\};$$

where $M_k$ and $N_k$ are the number of input arguments to $g_k(\cdot)$ and $g_k\{\cdot\}$, respectively, and $t_{k,i}$ is a formula in the language in which $x$ may appear. As in BLOG, each $f_k(x)$ is associated with a random variable $\phi_{f_k}[x]$. The random variables $\phi_{g_1}, \ldots, \phi_{g_K}$, including all those implicated in the terms, must have been generated by other lines or are observed. Overloading the notation, we define the terms $t_{k,i}$ to be random variables that depend deterministically on other generated or observed random variables. The set of statements (7) defines the generative process

$$\pi_v \sim \text{Stick}(\alpha_v) \qquad (8)$$
$$\phi_{f_k}[x] \sim \phi_{g_k}[t_{k,1}, \ldots, t_{k,M_k}](t_{k,M_k+1}, \ldots, t_{k,M_k+N_k}), \qquad (9)$$

for $k = 1, \ldots, K$, $x = 1, \ldots, \infty$, where $\alpha_v$ is the user-defined DP concentration parameter and $\pi_v$ is a multinomial distribution such that each success rate parameter $\pi_{v,x}$ determines the probability of choosing a particular object $x$. NP-BLOG infers a distribution $\pi$ over objects of type $v$ following the condition of infinite exchangeability. For example, applying rules (8,9) to line 4 of Fig. 3, we get

$$\pi_{\text{Author}} \sim \text{Stick}(\alpha_{\text{Author}})$$
$$\phi_{\text{Name}}[a] \sim \phi_{\text{NameDist}}, \quad \text{for } a = 1, \ldots, \infty$$

If an object type does not have any dependency or number statements, then no distribution over its extension is introduced (e.g. strings in the citation matching model).

The implementation of the DP brings about an important subtlety: if $x$ takes on a possibly infinite different set of values, how do we recover the true number of objects $n(\tau)$? The idea is to introduce a bijection from the subset of positive natural numbers that consists only of *active* objects to the set $\{1, \ldots, n(\tau)\}$. An object is active in possible world $\omega$ if and only if at least one random variable is assigned to that object in $\omega$. In the above example, $n(\text{Author})$ is the number of author objects that are mentioned in the citations. Of course, in practice we do not sample an infinite series of random variables $\phi_{\text{Name}}[a]$.

If we declare a function symbol $f$ with a return type $\tau$ ranging over a set of unknown objects, then there exists the default generating process

$$z_f[x] \sim \pi_\tau. \qquad (10)$$

We use $z_f[x]$ instead of $\phi_f[x]$ to show that the random variables are the indicators of the DP mixture (4). For example, each $z_{\text{RefPub}}[c]$ in line 6 in Fig. 3 is independently drawn from the distribution of publications $\pi_{\text{Pub}}$. We can view line 6 as constructing a portion of the hierarchical model, as shown in Fig. 4. The number of publications $n(\text{Pub})$ is set to the number of different values assigned to $z_{\text{RefPub}}[c]$.

NP-BLOG allows for the definition of a symbol $f$ that corresponds to a multinomial distribution over $[\tau]$, so its range is $\mathcal{M}_{\text{Multinomial}}([\tau])$. It exhibits the default prior

$$\phi_f[x] \sim \text{Dirichlet}(\alpha_f \pi_\tau), \qquad (11)$$

analogous to (10). $\alpha_f$ is a user-defined scalar. We define $n_f[x]$ to be the true number of objects associated with collection $f(x)$. This is useful for modeling collections of objects such as the authors of a publication. Applying rules (8,9,11) to the statements in Fig. 3 involving publication objects, we arrive at the generative process

$$\pi_{\text{Pub}} \sim \text{Stick}(\alpha_{\text{Pub}})$$
$$\phi_{\text{Title}}[p] \sim \phi_{\text{TitleDist}}, \quad \text{for } p = 1, \ldots, n(\text{Pub})$$
$$\phi_{\text{PubAuthorsDist}}[p] \sim \text{Dirichlet}(\alpha_{\text{PubAuthorsDist}} \pi_{\text{Author}}).$$

Most of the corresponding graphical model is shown in Fig. 4. Only the $\phi_{\text{PubAuthorsDist}}[p]$'s are missing, and they are shown in Fig. 5. The true number of authors $n_{\text{PubAuthorsDist}}[p]$ in publication $p$ comes from the support of all random variables that refer to it, and $n(\text{Pub})$ is determined by $n_{\text{PubAuthorsDist}}$. While this paper focuses on the Dirichlet process, our framework allows for other classes of nonparametric distributions. One example can be found in the aircraft tracking domain from Sec. 3.2, in which the generation of aircraft transition tables might be specified with the statement StateTransDist(a) $\sim$ StateTransPrior{}.

In both cases (10) and (11), one can override the defaults by including appropriate dependency statements for $f$, in which case we get $\phi_f[x] \sim \phi_g$, following rule (9). For example, lines 7 and 8 in Fig. 3 specify the generative process for the author mention objects,

$$z_{\mathsf{RefAuthor}}[u] \sim \phi_{\mathsf{PubAuthorsDist}}[p]$$
$$\phi_{\mathsf{CitedName}}[u] \sim \phi_{\mathsf{NameStrDist}}(\phi_{\mathsf{Name}}[a]) ,$$
$$\text{s.t. } p = z_{\mathsf{RefPub}}[c],\ c = \phi_{\mathsf{CitedIn}}[u],\ a = z_{\mathsf{RefAuthor}}[u].$$

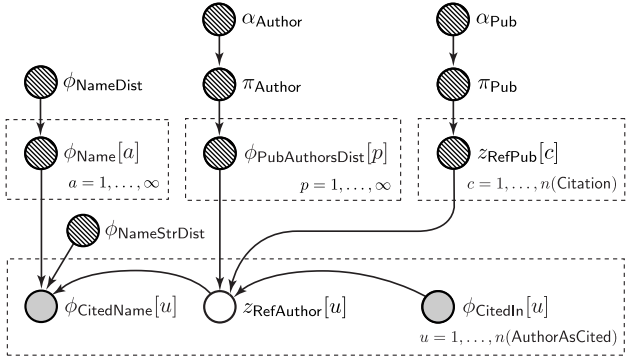

Figure 5: The white nodes are the portion of the graphical model generated in lines 7 and 8 of Fig. 3. See Fig. 4 for an explanation of the darkened nodes.

| | Face | Reinforce. | Reason. | Constraint |
|---|---|---|---|---|
| Num. citations | 349 | 406 | 514 | 295 |
| Num. papers | 246 | 149 | 301 | 204 |
| Phrase matching | 0.94 | 0.79 | 0.86 | 0.89 |
| RPM+MCMC | 0.97 | 0.94 | 0.96 | 0.93 |
| CRF-Seg ($N=9$) | 0.97 | 0.94 | 0.94 | 0.95 |
| NP-BLOG | 0.93 | 0.84 | 0.89 | 0.86 |

Table 1: Citation matching results for the Phrase Matching [9], RPM [14], CRF-Seg [22] and NP-BLOG models. Performance is measured by counting the number of publication clusters that are recovered perfectly. The NP-BLOG column reports an average over 1000 samples.

Fig. 5 shows the equivalent graphical model.

The generative process (8,9) is a stick-breaking construction over the unknown objects and their attributes. When the objects $x$ range over the set of natural numbers, (8,9) is equivalent to the Dirichlet process

$$G_v \sim DP\left(\alpha_v, H_{v,1} \times \cdots \times H_{v,K}\right), \qquad (12)$$

where $G_v \triangleq \sum_{x=1}^{\infty} \pi_{v,x} \delta(\phi_{f_1}[x]) \times \cdots \times \delta(\phi_{f_K}[x])$, and $H_{v,k}$ is the base measure over the assignments to $\phi_{f_k}$, defined by $g_k$ conditioned on the terms $t_{k,1}, \ldots, t_{k,M_k+N_k}$.

Since BLOG is a typed, free language, we need to allow for the null assignment to $\phi_f[x]$ when it is implicitly drawn from $\pi_\tau$ in (10). We permit the clause

$$f(x) \sim \text{if } cond \text{ then null}; \qquad (13)$$

which defines $\phi_f[x] \sim \delta(\mathsf{null})\delta(cond) + \pi_\tau(1-\delta(cond))$. This statement is necessary to take care of the situation when an object's source can be of different types, as in the aircraft tracking domain with false alarms [10].

Next, we briefly describe how to extend the rules of semantics to functions with multiple input arguments. Let's consider the case of two inputs with an additional logical variable $y \in [\nu]$. Handling an additional input argument associated with known (guaranteed) objects is easy. We just duplicate (8,9) for every instance of $y$ in the guaranteed type extension. This is equivalent to adding a finite series of plates in the graphical model. Otherwise, we assume the unknown objects are drawn independently. That is, $\pi_{(v,\nu)} = \pi_v \pi_\nu$. Multiple unknown objects as input does cause some superficial complications with the interpretation of (8,9) as a DP, principally because we need to define new notation for products of measures over different types.

The DP determines an implicit distribution of unknown, infinitely exchangeable objects according to their properties. That is, the DP distinguishes unknown objects solely by their attributes. However, this is not always desirable — for instance, despite being unable to differentiate the individual pieces, we know a chess board always has eight black pawns. This is precisely why we retain the original number statement syntax of BLOG which allows the user to specify a prior over the number of unknown objects, independent of their properties. In the future, we would like to experiment with priors that straddle these two extremes. This could possibly be accomplished by setting a prior on the Dirichlet concentration parameter, $\alpha$.

By tracing the rules of semantics, one should see that only thing the citation matching model does not generate is values for $\mathsf{CitedIn}(u)$. Therefore, they must be observed. We can also provide observations from any number of object attributes, such as $\mathsf{CitedTitle}(c)$ and $\mathsf{CitedName}(u)$, which would result in unsupervised learning. By modifying the set of evidence, one can also achieve supervised or semi-supervised learning. Moreover, the language can capture both generative and discriminative models, depending whether or not the observations are generated.

To summarize, the rules given by (7-11,13), combined with the number statement [11], construct a distribution $p(\phi, z, n, \gamma)$ such that the set of auxiliary variables is $\gamma = \{\pi, \alpha\}$, $\{\phi, z\}$ is in one-to-one correspondence with the interpretations of the function symbols, the $n$ are the sizes of the $[\tau]$, and an assignment to $\{\phi, z, n\}$ completely determines the possible world $\omega \in \Omega$. The rules of semantics assemble models that are arbitrary hierarchies of DPs.

## 4 Experiment

The purpose of this experiment is to show that the NP-BLOG language we have described realizes probabilistic inference on a real-world problem. We simulate the citation matching model in Fig. 3 on the CiteSeer data set [9], which consists of manually segmented citations from four research areas in AI.

We use Markov Chain Monte Carlo (MCMC) to simulate possible worlds from the model posterior given evidence in the form of cited authors and titles. Sec. 2 briefly describes the inference engine. Table 1 compares the performance of the NP-BLOG model to [14, 22] and the greedy agglomerative clustering method [9] (implemented by [14]). We achieve respectable matching accuracy, even though the specification of the model requires only a few lines in NP-BLOG, implements naive string metrics (the Jaro metric for author surnames and the standard TF-IDF information retrieval metric for distances between titles), and requires no supervision. By comparison, [22] uses as many as 9 different citation fields, and [14] tunes the parameters of the distributions using an alternate, labeled data set. (These richer attributes, with the exception of the random field implemented in [22], could be incorporated into NP-BLOG.) Since the segmented data is not publicly available, the our version of the data might be slightly different. Our NP-BLOG Gibbs sampling scheme is efficient; we found 1000 iterations was more than sufficient for data sets with as many as 500 exemplars. We caution, however, that determining a suitable stopping point for the Markov chain is, at the present state of research, more an art than a science.

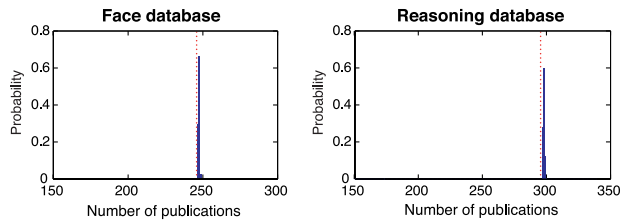

Figure 6: Estimated (solid blue) and true (dashed red line) number of publications for the Face and Reasoning data.

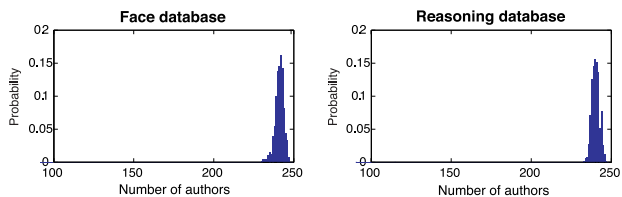

Figure 7: Estimated distribution of the hidden number of authors for the Face and Reasoning data sets.

In Figures 6 and 7, we plot the Monte Carlo estimate of the numbers of publication and author clusters for two data sets. The posteriors over the number of publications are highly peaked, and they closely match the ground truth.

## 5 Conclusions

This paper presented novel semantics for modeling collections of objects and their properties in arbitrary hierarchies by extending the BLOG probabilistic first-order language. We demonstrated that NP-BLOG models complex domains while concealing many implementation details from the user. We adopted Bayesian nonparametric methods, and notably Dirichlet processes, for defining distributions over collections of infinitely exchangeable unknown objects and their properties. Significantly, Dirichlet processes cohesively and efficiently handle model selection of unbounded sets of objects in first-order probabilistic inference.

There is much future work on this topic. An important direction is the development of efficient, flexible and on-line inference methods for hierarchies of Dirichlet processes.

**Acknowledgements**

This paper wouldn't have happened without the help of Brian Milch. Special thanks to Gareth Peters and Mike Klaas for their assistance, and to the reviewers for their time and effort in providing us with constructive comments.